\documentclass[sigconf,table]{acmart}

\usepackage{xcolor}
\usepackage{booktabs}
\usepackage{amsmath}
\usepackage{newtxtext,newtxmath}

\definecolor{top1}{RGB}{255, 223, 186}  % Light orange for 1st
\definecolor{top2}{RGB}{204, 255, 229}  % Light green for 2nd
\definecolor{top3}{RGB}{204, 229, 255}  % Light blue for 3rd

\begin{document}

\title{Flow Matters: Directional and Expressive GNNs for Heterophilic Graphs}

% \author{Anonymous}
% \affiliation{%

\author{Arman Gupta}
\affiliation{%
  \institution{Mastercard}
  % \city{Hekla}
  \country{India}}
\email{arman.gupta@mastercard.com}

\author{Govind Waghmare}
\affiliation{%
  \institution{Mastercard}
  % \city{Hekla}
  \country{India}}
\email{govind.waghmare@mastercard.com}

\author{Gaurav Oberoi}
\affiliation{%
  \institution{Mastercard}
  % \city{Hekla}
  \country{India}}
\email{gaurav.oberoi@mastercard.com}

\author{Nitish  Srivastava}
\affiliation{%
  \institution{Mastercard}
  % \city{Hekla}
  \country{India}}
\email{nitish.srivastava@mastercard.com}

% \include{abstract}

\begin{abstract}

In heterophilic graphs, where neighboring nodes often belong to different classes, conventional Graph Neural Networks (GNNs) struggle due to their reliance on local homophilous neighborhoods. Prior studies suggest that modeling edge directionality in such graphs can increase effective homophily and improve classification performance. Simultaneously, recent work on polynomially expressive GNNs shows promise in capturing higher-order interactions among features. In this work, we study the combined effect of edge directionality and expressive message passing on node classification in heterophilic graphs. Specifically, we propose two architectures: (1) a polynomially expressive GAT baseline (Poly), and (2) a direction-aware variant (Dir-Poly) that separately aggregates incoming and outgoing edges. Both models are designed to learn permutation-equivariant high-degree polynomials over input features, while remaining scalable with no added time complexity. Experiments on five benchmark heterophilic datasets show that our Poly model consistently outperforms existing baselines, and that Dir-Poly offers additional gains on graphs with inherent directionality (e.g., Roman Empire), achieving state-of-the-art results. Interestingly, on undirected graphs, introducing artificial directionality does not always help, suggesting that the benefit of directional message passing is context-dependent. Our findings highlight the complementary roles of edge direction and expressive feature modeling in heterophilic graph learning.
\end{abstract}




\maketitle

\section{Introduction}
GNNs have become the de facto approach for learning over graph-structured data, achieving remarkable success in domains ranging from social network analysis and recommendation systems to molecular property prediction and fraud detection \cite{Comprehensive_Survey_on_GNNs_@Wu}. However, the majority of GNN architectures \cite{GCN@kipf, graphattentionnetworks@veličković,GraphSage@Hamilton} assume that the input graphs are undirected and homophilic, that is, neighboring nodes tend to belong to the same class or share similar attributes. This assumption underpins the core message-passing mechanism of many GNNs, where information is aggregated from local neighbors to infer node properties.

Yet, many real-world graphs such as citation networks, web graphs, and interaction graphs, are inherently directed and heterophilic. In heterophilic graphs, connected nodes frequently belong to different classes, violating the inductive bias of classical GNNs \cite{GNN_Beyond_Homophily@Zhu}. Moreover, transforming directed graphs into undirected ones, often done during preprocessing (e.g., in libraries like PyTorch Geometric \footnote{\url{https://github.com/pyg-team/pytorch_geometric/blob/master/torch_geometric/datasets/heterophilous_graph_dataset.py##L119}}), may lead to information loss, especially in heterophilic settings. Recent studies \cite{edgedirectionalityimproveslearning@rossi2023} have shown that preserving edge directionality can increase the effective homophily of a graph and substantially improve GNN performance on node classification tasks in such settings.
At the same time, another critical limitation of most existing GNN architectures is their limited expressive power. Traditional GNNs like GCN \cite{GCN@kipf}, GraphSAGE \cite{GraphSage@Hamilton}, or GAT \cite{graphattentionnetworks@veličković} rely on simple, linear aggregation functions (e.g., sum, mean, and max) which cannot capture high-order, nonlinear feature interactions among nodes. This expressivity bottleneck has prompted the development of more flexible aggregation functions such as principled neighborhood aggregation (PNA) \cite{PNA@Corso} by combining multiple simple aggregation functions, each associated with a learnable weight but the practical capacity of PNA is still limited by simply combining multiple simple aggregation functions. Recent literature have introduced higher-order polynomially expressive models such as tGNN \cite{tGNNhighorderpoolinggraphneural@Hua} and Polynormer\cite{polynormerpolynomialexpressivegraphtransformer@deng2024}, which aim to represent richer feature transformations by learning high-degree polynomial functions while preserving permutation equivariance. However, these models either suffer from limited polynomial expressivity \cite{tGNNhighorderpoolinggraphneural@Hua} or incur high computational costs \cite{dhilgtscalablegraphtransformer@liao2024}, making them less practical for large-scale or complex graph tasks.

Although both edge directionality and polynomial expressiveness have independently been shown to improve performance in specific contexts, their combined effect on heterophilic node classification remains largely unexplored. In this work, we bridge this gap by empirically studying the synergistic impact of edge directionality and expressive message passing on heterophilic graphs. Specifically, we design two architectures:
\begin{itemize}
    \item \textbf{Poly}, a GAT-based model augmented with a permutation-equivariant polynomially expressive message passing scheme. Inspired by the local attention mechanism in Polynormer~\cite{polynormerpolynomialexpressivegraphtransformer@deng2024}, Poly explicitly captures high-degree interactions by recursively composing local attention and linear layers with multiplicative gating, and
    \item \textbf{Dir-Poly}, which additionally accounts for edge direction by performing separate aggregations over incoming and outgoing edges.
\end{itemize}

Both models are scalable and do not incur additional time complexity compared to standard GATs. We evaluate these architectures on five benchmark heterophilic datasets: Roman Empire, Amazon Ratings, Minesweeper, Tolokers, and Questions. Our results show that Poly consistently outperforms all existing baselines, while Dir-Poly achieves further gains on inherently directed graphs like Roman Empire. However, enforcing artificial directionality on undirected graphs offers limited benefit, indicating that direction-aware message passing is context-dependent. Overall, our results show that directional message passing and high-degree polynomial aggregation are complementary, and combining them effectively models complex interactions in heterophilic graphs.
% Our findings suggest that directional message passing and high-degree polynomial aggregation are complementary and that jointly leveraging them is an effective strategy for modelling complex node interactions in heterophilic graphs.

% Both models are designed to be scalable and efficient and do not incur additional time complexity compared to standard GATs. We evaluate these architectures on five benchmark heterophilic datasets: Roman Empire, Amazon Ratings, Minesweeper, Tolokers, and Questions. Our results show that Poly outperforms all existing baselines, and Dir-Poly further improves performance on inherently directed graphs, such as Roman Empire. Interestingly, when applied to graphs without natural directionality, artificially enforcing edge directionality does not always yield gains—suggesting that the benefits of direction-aware message passing are context-dependent.

% Our findings suggest that directional message passing and high-degree polynomial aggregation are complementary and that jointly leveraging them is an effective strategy for modelling complex node interactions in heterophilic graphs.

\section{Related Work}

Graph Neural Networks (GNNs) have emerged as core models for learning from graph-structured data. Early frameworks such as GCN~\cite{GCN@kipf}, GAT~\cite{graphattentionnetworks@veličković}, and GraphSAGE~\cite{GraphSage@Hamilton} established neighborhood aggregation as the fundamental paradigm, typically using permutation-invariant functions like sum or mean. While effective on homophilic graphs, where neighboring nodes often share labels, these models degrade in performance on heterophilic graphs due to their implicit homophily assumptions.

\textbf{GNNs for Heterophily.}
To address this, recent models such as H2GCN~\cite{GNN_Beyond_Homophily@Zhu}, GPR-GNN~\cite{GRPGNN@Chien}, and GGCN~\cite{GGCNsidescoinheterophilyoversmoothing@yan2022} introduce mechanisms to disentangle structural proximity from label similarity, either by decoupling feature propagation or adopting flexible message-passing schemes. More recently, Dir-GNN~\cite{edgedirectionalityimproveslearning@rossi2023} demonstrated that explicitly modeling edge directionality improves performance on heterophilic datasets by distinguishing between incoming and outgoing flows, an insight especially relevant for real-world directed graphs like citation or interaction networks.

\textbf{Directionality in GNNs.}
Although several earlier GNNs~\cite{neuralmessagepassingquantum@gilmer2017,TheGraphNeuralNetworkModel@2009,gatedgraphsequenceneural@li2017} nominally support directed graphs, many implementations symmetrize edges or only partially utilize directional information (e.g., GatedGCN~\cite{gatedgraphsequenceneural@li2017} aggregates solely from out-neighbors). Spectral methods like DGCN~\cite{directedgraphconvolutionalnetwork@tong2020}, DiGCN~\cite{digraph@tong2020}, and MagNet~\cite{magnetneuralnetworkdirected@zhang2021} incorporate directional operators, but often share weights across directions and suffer from limited scalability due to computationally expensive spectral decompositions. Dir-GNN~\cite{edgedirectionalityimproveslearning@rossi2023} overcomes these limitations via spatial aggregation paths that preserve direction-specific semantics, showing notable gains on heterophilic tasks.

\textbf{Expressive Aggregation Functions.}
Parallel to directionality, another line of work enhances GNN expressivity through advanced aggregation. PNA~\cite{PNA@Corso} improves flexibility using multiple aggregators and degree-scalers. Polynormer~\cite{polynormerpolynomialexpressivegraphtransformer@deng2024} further introduces polynomially expressive attention to model high-order interactions while preserving permutation equivariance. Tensor-based approaches~\cite{tGNNhighorderpoolinggraphneural@Hua} similarly aim to capture higher-order structures, albeit with high computational cost. These methods underscore the importance of moving beyond basic aggregators to capture richer structural patterns, but often face trade-offs in scalability or expressivity.

\textbf{Joint Modeling of Direction and Expressivity.}
Despite progress in direction-aware and expressive GNNs, their integration remains limited. Our work bridges this gap by introducing a unified framework that combines directional message passing with polynomial aggregation. Specifically, we extend GAT by (1) introducing permutation-equivariant polynomial aggregators and (2) modeling directed edge flow via separate incoming and outgoing paths. Unlike prior methods that trade efficiency for performance, our approach remains computationally tractable while achieving state-of-the-art results on multiple heterophilic benchmarks.

\begin{table*}[t]
\centering
\caption{Averaged node classification results over 10 runs on heterophilic datasets — Accuracy is reported for \textit{roman-empire} and \textit{amazon-ratings}, and ROC AUC for the rest. We highlight the top-3 scores in each column using \colorbox{top1}{orange (1st)}, \colorbox{top2}{green (2nd)}, and \colorbox{top3}{blue (3rd)} for improved visual comparison.}
\label{tab:main-results}
% \scriptsize
\begin{tabular}{lccccc}
\toprule
\textbf{Model} & \textbf{roman-empire} & \textbf{amazon-ratings} & \textbf{minesweeper} & \textbf{tolokers} & \textbf{questions} \\
\midrule
GCN & 73.69 ± 0.74 & 48.70 ± 0.63 & 89.75 ± 0.52 & 83.64 ± 0.67 & 76.09 ± 1.27 \\
GraphSAGE & 85.74 ± 0.67 & \cellcolor{top1}53.63 ± 0.39 & 93.51 ± 0.57 & 82.43 ± 0.44 & 76.44 ± 0.62 \\
GAT-sep & 88.75 ± 0.41 & 52.70 ± 0.62 & \cellcolor{top2}93.91 ± 0.35 &  \cellcolor{top3}83.78 ± 0.43 & 76.79 ± 0.71 \\
H2GCN & 60.11 ± 0.52 & 36.47 ± 0.23 & 89.71 ± 0.31 & 73.35 ± 1.01 & 63.59 ± 1.46 \\
GPRGNN & 64.85 ± 0.27 & 44.88 ± 0.34 & 86.24 ± 0.61 & 72.94 ± 0.97 & 55.48 ± 0.91 \\
FSGNN & 79.92 ± 0.56 & 52.74 ± 0.83 & 90.08 ± 0.70 & 82.76 ± 0.61 & \cellcolor{top1}78.86 ± 0.92 \\
GloGNN & 59.63 ± 0.69 & 36.89 ± 0.14 & 51.08 ± 1.23 & 73.39 ± 1.17 & 65.74 ± 1.19 \\
GGCN & 74.46 ± 0.54 & 43.00 ± 0.32 & 87.54 ± 1.22 & 77.31 ± 1.14 & 71.10 ± 1.57 \\
OrderedGNN & 77.68 ± 0.39 & 47.29 ± 0.65 & 80.58 ± 1.08 & 75.60 ± 1.36 & 75.09 ± 1.00 \\
G2-GNN & 82.16 ± 0.78 & 47.93 ± 0.58 & 91.83 ± 0.56 & 82.51 ± 0.80 & 74.82 ± 0.92 \\
DIR-GNN & \cellcolor{top3}91.23 ± 0.32 & 47.89 ± 0.39 & 87.05 ± 0.69 & 81.19 ± 1.05 & 76.13 ± 1.24 \\
tGNN & 79.95 ± 0.75 & 48.21 ± 0.53 & 91.93 ± 0.77 & 70.84 ± 1.75 & 76.38 ± 1.79 \\
\midrule
\textbf{Poly (Ours)} & \cellcolor{top2}91.57 ± 0.46 & \cellcolor{top2}53.50 ± 0.39 & \cellcolor{top1}95.32 ± 0.28 & \cellcolor{top1}84.87 ± 0.76 & \cellcolor{top2}77.31 ± 0.87 \\
\textbf{Dir-Poly (Ours)} & \cellcolor{top1}94.51 ± 0.22 & \cellcolor{top3}50.73 ± 0.56 & \cellcolor{top3}93.74 ± 0.70 & \cellcolor{top2}84.10 ± 0.74 & \cellcolor{top3}76.87 ± 1.04 \\
\bottomrule
\end{tabular}
\end{table*}

\section{Background}

Graph Neural Networks (GNNs) have become the cornerstone of learning on graph-structured data. However, their effectiveness often hinges on the nature of the graph topology, particularly, the degree of \textit{homophily} \cite{edgedirectionalityimproveslearning@rossi2023} and whether \textit{edge directions} are available and informative.
\subsection*{Notation}
Let $G = (V, E)$ denote a graph, where $V = \{v_i\}_{i=1}^N$ is the set of nodes and $E$ is the set of edges. The adjacency matrix of the graph is represented as $A \in \{0, 1\}^{N \times N}$, where $A_{ij} = 1$ if $(i, j) \in E$, and $0$ otherwise. Let $X \in \mathbb{R}^{N \times d}$ denote the node feature matrix, where $x_i$ is the $d$-dimensional feature vector for node $v_i$.

The variables used in the equations are as follows:
\begin{itemize}
    \item $x^{(i)}$: Feature matrix at the $i$-th layer.
    \item $x^{(i-1)}$: Input feature matrix from the $(i-1)$-th layer.
    \item $W_{\text{in}}$: Learnable input transformation matrix.
    \item $W^{(h)}_i$: Weight matrix for nonlinear transformation at layer $i$.
    \item $W^{(l)}_i$: Linear transformation weight matrix at layer $i$.
    \item $h^{(i)}$: Nonlinear activation of features at layer $i$.
    \item $\text{Conv}_i(\cdot)$: Graph convolution or attention operator at layer $i$.
    \item $\beta_i$: Learnable coefficient controlling interpolation between nonlinear and linear components.
    \item $\odot$: Elementwise (Hadamard) product.
    \item $\sigma(\cdot)$: Nonlinear activation function (e.g., ReLU, sigmoid).
    \item $x_{\text{local}}$: Output feature after summing across all $L$ layers.
    \item $m^{(k)}_{i,\leftarrow}$: Message from in-neighbors at layer $k$.
    \item $m^{(k)}_{i,\rightarrow}$: Message from out-neighbors at layer $k$.
    \item $\text{AGG}^{(k)}_{\leftarrow}, \text{AGG}^{(k)}_{\rightarrow}$: Aggregators for in- and out-neighbors at layer $k$.
    \item $\text{COM}^{(k)}$: Combination function aggregating messages and current node feature at layer $k$.
\end{itemize}

\subsection{Homophily and Heterophily in Graphs}

Most Graph Neural Networks (GNNs) rely on the \textit{homophily assumption}—that neighboring nodes tend to share the same label. While valid for many graphs (e.g., citation networks), this assumption breaks down in \textit{heterophilic} settings such as fraud detection or social networks, where connected nodes may differ significantly in labels. To measure homophily, the \textbf{node homophily} metric is used:
\begin{equation}
    h = \frac{1}{|V|} \sum_{i \in V} \frac{1}{d_i} \sum_{j: (i,j) \in E} \mathbb{I}[y_i = y_j]
\end{equation}
where $\mathbb{I}[y_i = y_j]$ is 1 if nodes $i$ and $j$ share the same label, and 0 otherwise.

\noindent For directed or weighted graphs, the \textbf{weighted node homophily} generalizes this:
\begin{equation}
    h(S) = \frac{1}{|V|} \sum_{i \in V} \frac{\sum_{j} s_{ij} \mathbb{I}[y_i = y_j]}{\sum_{j} s_{ij}}
\end{equation}
where $S$ is a message-passing matrix (e.g., adjacency $A$, $A^\top$, or multi-hop variants). These metrics reveal the need for models that go beyond local label similarity to succeed in heterophilic graphs.

\subsection{Polynomial Expressivity in Message Passing GNNs}

Traditional message passing neural networks (MPNNs) follow a linear aggregation scheme:
\begin{equation}
    P(X)_i = \sum_j c_{i,j} X_j
\end{equation}
where $c_{i,j}$ is a learnable or pre-defined edge weight. Since the output is a \textit{linear combination} of neighbor features, these models are \textit{at most 1-polynomial expressive} \cite{tGNNhighorderpoolinggraphneural@Hua, polynormerpolynomialexpressivegraphtransformer@deng2024}, relying on activation functions to introduce nonlinearity. This motivates models capable of capturing higher-degree interactions explicitly.

\subsection{Edge Directionality in Graphs}

% Another limitation of standard GNNs is the neglect of edge directionality, crucial in domains like knowledge graphs and citation networks. Inspired by the Directed GNN (Dir-GNN) framework \cite{edgedirectionalityimproveslearning@rossi2023}, Dir-Poly independently aggregates information from both in- and out-neighbors:
Another limitation of standard GNNs is that they often neglect edge directionality, which is crucial in domains such as knowledge graphs and citation networks. Inspired by the Directed GNN (Dir-GNN) framework~\cite{edgedirectionalityimproveslearning@rossi2023}, our Dir-Poly model addresses this by independently aggregating information from both incoming and outgoing neighbors:
\begin{align}
    m^{(k)}_{i,\leftarrow} &= \text{AGG}^{(k)}_{\leftarrow}\left\{ (x^{(k-1)}_j, x^{(k-1)}_i) : (j, i) \in E \right\} \\
    m^{(k)}_{i,\rightarrow} &= \text{AGG}^{(k)}_{\rightarrow}\left\{ (x^{(k-1)}_j, x^{(k-1)}_i) : (i, j) \in E \right\} \\
    x^{(k)}_i &= \text{COM}^{(k)}\left(x^{(k-1)}_i, m^{(k)}_{i,\leftarrow}, m^{(k)}_{i,\rightarrow} \right)
\end{align}
This directional formulation enhances both expressivity and performance, especially on heterophilic graphs.

\section{Methodology}

We propose two complementary models, \textbf{Poly} and \textbf{Dir-Poly}, to overcome the expressivity and directionality limitations of standard GNNs.

\subsection{Polynomial Attention Model (Poly)}

Poly increases expressivity by leveraging attention mechanisms, enabling the learning of polynomial transformations.
\begin{equation}
    x_{\text{in}} = W_{\text{in}} x
\end{equation}

\begin{align}
    \text{for } i &= 1 \text{ to } L:
    \begin{cases}
        h^{(i)} &= \sigma(W^{(h)}_i x^{(i-1)}) \\
        x' &= \text{Conv}_i(x^{(i-1)}) + W^{(l)}_i x^{(i-1)} \\
        x^{(i)} &= (1 - \beta_i) \cdot (h^{(i)} \odot x') + \beta_i \cdot x'
    \end{cases} 
\end{align}
\begin{equation}
    x_{\text{local}} = \sum_{i=1}^{L} x^{(i)}
\end{equation}
Here, \texttt{Conv}$_i$ is a graph attention convolution (e.g., GAT), and \(\sigma\) is a nonlinearity. The learnable coefficients \(\beta_i\) control the interpolation between attention-based and linear contributions.

\subsubsection*{Proof of Polynomial Expressivity}
To prove that Poly is polynomially expressive, consider the recurrence relation:
\begin{multline}
    x^{(i)} = (1 - \beta_i) \left(\sigma(W^{(h)}_i x^{(i-1)}) \odot (\text{Conv}_i(x^{(i-1)}) + W^{(l)}_i x^{(i-1)})\right) \\
    + \beta_i \left(\text{Conv}_i(x^{(i-1)}) + W^{(l)}_i x^{(i-1)}\right)
\end{multline}
At each layer $i$, this equation composes previous features $x^{(i-1)}$ using:
\begin{itemize}
    \item a learned nonlinearity $\sigma$ (e.g., ReLU or sigmoid),
    \item an attention-based aggregation (\texttt{Conv}$_i$), and
    \item multiplicative interactions (via elementwise product $\odot$).
\end{itemize}
The result is that $x^{(i)}$ is a polynomial function of $x$ with degree increasing in $i$:
\begin{equation}
    \deg(x^{(i)}) \leq i
\end{equation}
This is because each $x^{(i)}$ is built from elementwise and linear operations over $x^{(i-1)}$, which themselves are polynomials of degree $i-1$. By stacking $L$ layers, $x_{\text{local}} = \sum_{i=1}^{L} x^{(i)}$ becomes a sum of polynomials up to degree $L$. Therefore, Poly is \textbf{L-polynomial expressive}.

\subsection{Directed Polynomial Model (Dir-Poly)}

Dir-Poly extends Poly by instantiating \texttt{Conv}$_i$ as a Directed GAT, separately handling in- and out-neighbors. The remaining architecture remains unchanged, preserving polynomial expressivity while incorporating directional inductive bias.

\subsection{Implications}
\noindent \textbf{Poly} models high-degree polynomial interactions using local attention, outperforming shallow MPNNs.
\noindent \textbf{Dir-Poly} introduces direction-aware message passing, further improving performance on asymmetric and heterophilic graphs.
\noindent Together, they enable robust learning on diverse graph structures.

\section{Experiments}
\subsection{Datasets and Metrics}
We evaluate our proposed models, \textbf{Poly} and \textbf{Dir-Poly}, on five benchmark heterophilic datasets \cite{criticallookevaluationgnns@platonov2024}:  \textit{minesweeper}, \textit{tolokers}, \textit{questions}, \textit{roman-empire} and \textit{amazon-ratings}. For \textit{roman-empire} and \textit{amazon-ratings}, we report average node classification \textbf{Accuracy}, while for the remaining three datasets, \textit{minesweeper}, \textit{tolokers}, and \textit{questions}, we use the \textbf{ROC AUC} score. All results are averaged over 10 random runs with different seeds.

\subsection{Baselines}

We compare our models, \textbf{Poly} and \textbf{Dir-Poly}, against several prominent GNN architectures, spanning both homophilic and heterophilic graph learning paradigms:

\noindent \textbf{Homophilic GNNs:} Models designed for graphs where connected nodes typically share similar labels. Examples include GCN~\cite{GCN@kipf}, GraphSAGE~\cite{GraphSage@Hamilton}, and GAT-sep~\cite{graphattentionnetworks@veličković}.

\noindent \textbf{Heterophilic GNNs:} Models that perform well even when neighboring nodes have dissimilar labels. Notable methods include H2-GCN ~\cite{GNN_Beyond_Homophily@Zhu}, GPRGNN~\cite{GRPGNN@Chien}, FSGNN~\cite{FSGNN@maurya2021}, Glo-GNN~\cite{GloGNNfindingglobalhomophilygraph@li2022}, and GGCN~\cite{GGCNsidescoinheterophilyoversmoothing@yan2022}.

\noindent \textbf{Direction-aware and order-sensitive GNNs:} Models that incorporate edge direction, temporal order, or higher-order dependencies for better expressivity. Examples include OrderedGNN~\cite{orderedgnnorderingmessage@song2023}, G2-GNN~\cite{G2GNNgradientgatingdeepmultirate@rusch2023}, DIR-GNN~\cite{edgedirectionalityimproveslearning@rossi2023}, and tGNN~\cite{tGNNhighorderpoolinggraphneural@Hua}.

\subsection{Results and Analysis}

In this section, we report the average performance over 10 random runs across five benchmark datasets, as summarized in Table~\ref{tab:main-results}.  
% Table~\ref{tab:main-results} shows averaged results over 10 random runs.

\noindent \textbf{Scalability of Poly} \
Poly retains architectural simplicity and computational efficiency despite introducing polynomial expressivity in message passing. Built on top of GAT, it applies element-wise polynomial transformations to GAT convolutions, without significantly increasing the overall parameter count or runtime. This lightweight and flexible design allows Poly to scale effectively to large graphs, making it suitable for real-world applications where both expressivity and efficiency are essential.

\noindent \textbf{Performance of Poly} \
Poly consistently achieves strong and reliable performance across all datasets. It attains the best results on \textit{minesweeper} (95.32) and \textit{tolokers} (84.87), and secures top-2 ranks on \textit{roman-empire}, \textit{amazon-ratings}, and \textit{questions}. These results clearly validate the benefit of polynomial transformations in enhancing the representational power of GNNs for modeling complex relationships in heterophilic graphs.

\noindent \textbf{Effectiveness of Edge Directionality in Dir-Poly} \
Dir-Poly, which augments Poly with direction-aware message passing, achieves the best performance on \textit{roman-empire} (94.51), improving over Poly by nearly 3 points. It also ranks in the top three on all five datasets, including second-best performance on \textit{tolokers} and third-best on \textit{amazon-ratings}, \textit{minesweeper}, and \textit{questions}. This indicates that incorporating edge directionality can effectively capture asymmetrical dependencies in graphs where such structure is meaningful.

\noindent \textbf{When Directionality Helps—or Not} \
Dir-Poly is not always better than Poly, as Poly performs better on all datasets except \textit{roman-empire}. 
This suggests that directional aggregation is beneficial primarily when edge directions encode informative structure, and may even degrade performance  otherwise.

% \noindent \textbf{Consistent Top Performance} \
% Poly and Dir-Poly are the only methods to consistently rank in the top three across all datasets. This robustness highlights the generality of polynomial expressivity and directional reasoning in diverse graph types.

\noindent \textbf{Comparison to GAT} \
Both models, \textbf{Poly} and \textbf{Dir-Poly}, outperform the GAT baseline across the board. Poly improves over GAT by 1.41 points on \textit{minesweeper} and by over 1 point on \textit{tolokers}, confirming that polynomial attention significantly boosts GAT’s representational power in heterophilic graphs.

\section{Conclusion}
In this work, we investigated the joint impact of edge directionality and polynomially expressive message passing for node classification on heterophilic graphs. Our proposed Poly model, which augments a GAT backbone with high-degree polynomial aggregation, consistently outperforms standard and direction-aware GNN baselines across diverse datasets. The Dir-Poly variant achieves further gains on inherently directed graphs (e.g., Roman Empire), highlighting the benefit of directional information. However, its mixed performance on undirected graphs indicates that the advantage of direction-aware aggregation is context-dependent. These results suggest that expressivity and edge direction are complementary, and future GNNs can benefit from adaptively leveraging both based on graph structure.

\balance

\bibliographystyle{ACM-Reference-Format}
\bibliography{main}

\end{document}